# A New Learning Method for Inference Accuracy, Core Occupation, and Performance Co-optimization on TrueNorth Chip

[1] Wei Wen, [1] Chunpeng Wu, [1] Yandan Wang, [1] Kent Nixon, [2] Qing Wu, [2] Mark Barnell, [1] Hai Li, [1] Yiran Chen

[1] University of Pittsburgh, Pittsburgh, PA, USA; [2] Air Force Research Laboratory, Rome, NY, USA

[1]{wew57, chw127, yaw46, kwn2, hal66, yic52}@pitt.edu, [2]{qing.wu.2, Mark.Barnell.1}@us.af.mil

## ABSTRACT

*IBM TrueNorth chip uses digital spikes to perform neuromorphic computing and achieves ultrahigh execution parallelism and power efficiency. However, in TrueNorth chip, low quantization resolution of the synaptic weights and spikes significantly limits the inference (e.g., classification) accuracy of the deployed neural network model. Existing workaround, i.e., averaging the results over multiple copies instantiated in spatial and temporal domains, rapidly exhausts the hardware resources and slows down the computation. In this work, we propose a novel learning method on TrueNorth platform that constrains the random variance of each computation copy and reduces the number of needed copies. Compared to the existing learning method, our method can achieve up to 68.8% reduction of the required neuro-synaptic cores or 6.5× speedup, with even slightly improved inference accuracy.*

## 1. INTRODUCTION

Von Neumann architecture has achieved remarkable success in the history of computer industry. However, following the upscaling of design complexity and clock frequency of modern computer systems, *power wall* [1][2] and *memory wall* [3], which denote the high power density and low memory bandwidth [4][5] w.r.t. the fast on-chip data processing capability, become two major obstacles hindering the deployment of von Neumann architecture in data-intensive applications. *Neuromorphic computer*, therefore, starts to gain attentions in computer architecture society due to its extremely high data processing efficiency for cognitive applications: different from decoupling storage and computation as von Neumann architecture does, neuromorphic system directly maps the neural networks to distributed computing units each of which integrates both computing logic and memory [1][6][7].

In 2014 April, IBM unveiled *TrueNorth* - a new neuromorphic system composed of 4096 neuro-synaptic cores, 1 million neurons, and 256 million synapses. Memories ("synapses") and computing units ("neurons") of each neuro-synaptic core are placed in close proximity and the neuro-synaptic cores are connected through "axons" network. TrueNorth chip is extremely efficient in executing neural networks and machine learning algorithms and able to achieve a peak computation of 58 Giga synaptic operations per second at a power consumption of 145mW [1][8].

The high computing efficiency and parallelism of TrueNorth are achieved by sacrificing the precision of message stored and propagated in networks: inter-core communication is performed using binary spikes and synaptic weights are quantized by low-resolution integers [9]. It is contradictory to common practices of deep learning/neural networks that require some high precisions [10]. Directly deploying the learned model with a low precision on TrueNorth leads to inevitable inference accuracy degradation. To compensate this accuracy loss, the current workaround of TrueNorth is to average the results over multiple copies instantiated in spatial and temporal domains: In spatial domain, TrueNorth utilizes stochastic neural mode to mimic fractional synaptic weights [9]. The similar stochastic scheme can be also applied to temporal domain to generate spikes with fractional expected values. Besides the stochastic scheme, TrueNorth also supports many deterministic neural coding schemes (*i.e., rate code, population code, time-to-spike code* and *rank code* [9]) in temporal domain to represent non-binary message using multiple binary spikes. *Rate code*, for example, uses the number of the spikes occurring within a predefined number of time steps to indicate the amplitude of a signal [7]. However, this "official" workaround rapidly exhausts hardware resources when the network scale increases and a large number of spatial samples (copies) are instantiated. It also slows down the computation when many spike samples are generated.

In this work, we first formulate the standard TrueNorth learning and deploying method and study its compatibility to learning networks with floating-point precision; we then analyze the essential factors that affect the accuracy of TrueNorth during data quantization and task deployment; finally, we propose a new learning method to mitigate the adverse impacts of these factors on TrueNorth accuracy by constraining the variance of each copy through biasing the randomness of synaptic connectivity in neuro-synaptic cores. Experiments executed on real TrueNorth chip shows that our proposed scheme can achieve up to 68.8% reduction of the required neuro-synaptic cores or 6.5× performance speedup, with even slightly improved inference (*e.g.*, classification) accuracy.

## 2. PRELIMINARY

A TrueNorth chip is made of a network of neuro-synaptic cores. Each neuro-synaptic core includes a 256×256 configurable synaptic crossbar connecting 256 axons and 256 neurons in close proximity [7]. The inter-core communication is transmitted by spike events through axons. Determined by the axon type at each neuron, the weight of the corresponding synapse in the crossbar is selected from 4 possible integers.

As illustrated in Figure 1(a), in an artificial neural network, the output of a neuron ($z$) is determined by the input vector ($\mathbf{x}$) and the weights of the connections to the neuron ($\mathbf{w}$), such as

$$y = \mathbf{w} \cdot \mathbf{x} + b \text{, and} \quad (1)$$

$$z = h(y) . \quad (2)$$

Here $b$ is a bias. $h(\cdot)$ is a nonlinear activation function. The data in (1) and (2) usually are represented with floating-point precision.

When mapping the neural network to a neuro-synaptic core in TrueNorth, as shown in Figure 1(b), the input vector $\mathbf{x}$ is quantized into the spike format $\mathbf{x}'$ and the weights $\mathbf{w}$ are approximated by random synaptic weight samples **w'**. More specific, a pseudo random number generator (PRNG) is used to sample each connection in the synaptic crossbar: when the synapse is sampled to be "ON", the value indexed by the axon type will be applied as its weight; otherwise the synapse is disconnected. For the neuron implementation, rather than using nonlinear activation functions in traditional neural networks, TrueNorth adopts a spiking neuron model − *leaky integrate-and-fire* (LIF) model [9] to produce output in digital spike format. The model has 22 parameters (14 of which are user configurable) and 8 complex neuron specification equations. Regardless the possible versatility and complexity of LIF designs, a simple version of LIF, *e.g.*, McCulloch-Pitts neuron model, can be used to realize many high-level cognitive applications [11]. The operations of McCulloch-Pitts can be mathematically formulated as:

$$y' = \mathbf{w}' \cdot \mathbf{x}' - \lambda, \text{ and} \quad (3)$$

$$z' = \begin{cases} 1, \text{ reset } y'=0; \text{ if } y' \geq 0 \\ 0, \text{ reset } y'=0; \text{ if } y'<0 \end{cases}. \quad (4)$$

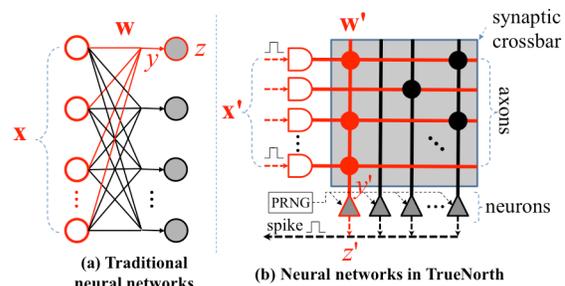

**Figure 1. Mapping neural networks in IBM TrueNorth.**

Here, $\lambda$ is a constant leak. Note that the spiking time is omitted in (3) and (4) because McCulloch-Pitts model is history-free.

The implementation of TrueNorth differs from traditional artificial neuron model in the following three aspects: First, the inputs (**x'**) and the neuron output (*z'*) are represented in a binary format of spikes; second, the weights (**w'**) are the integers selected from 4 synaptic weights, and the selections are determined by the axon type at the corresponding neurons [9]; finally, a digital and reconfigurable LIF model is used to represent the neuron function.

Such a digitized quantization operation inevitably results in system accuracy degradation. TrueNorth attempts to compensate this loss by introducing stochastic computation processes in temporal and spatial domains [9][11]: first, several neural codes are supported and used to provide a series of input spikes that follow the probabilities of the normalized input value (*e.g.*, pixel) intensities; second, the PRNG in LIF controls the synapse connectivity and makes the average value of the synaptic weight close to the one in the original neural network model with a high precision. As such, sufficient accuracy can be promised by averaging the computation results over all the spike samples in a long temporal window as well as a large number of the network instantiations of the neurosynaptic cores.

## 3. METHODOLOGY

However, temporal spike sampling method in TrueNorth significantly slows down the cognitive inference. In addition, the duplicated network copies will rapidly occupy the available cores as the scale of neural network model raises. In observation of these drawbacks of the TrueNorth solution, in this work, we propose *a synaptic connectivity probability-biased learning method* to speed up cognitive inference, reduce core occupation, and improve inference accuracy of the neural network deployed on TureNorth.

This section will provide the detailed theoretical explanation of our proposed learning method. We first formulate and analyze the learning method of TrueNorth, namely, *Tea learning* [11], and analyze its compatibility to the deep learning models. We then conduct a theoretical analysis of the expectation and variance of TrueNorth deployment. Based on this analysis, we propose our probability-biased learning method.

### 3.1 The Learning and Deploying of TrueNorth

Figure 2 gives an overview of the learning and deploying of a neural network on TrueNorth, which is named as *Tea learning method* by IBM [11]. Let's assume that McCulloch-Pitts neuron model is adopted. The corresponding stochastic operation on TrueNorth can be represented as follows:

$$y' = \sum_{i=0}^{n-1} w_i' x_i', \quad (5)$$

$$\begin{cases} P(w_i' = c_i) = p_i \\ P(w_i' = 0) = 1 - p_i \end{cases}, \quad (6)$$

$$p_i = w_i / c_i, \text{ and} \quad (7)$$

$$\begin{cases} P(x_i' = 1) = x_i \\ P(x_i' = 0) = 1 - x_i \end{cases}. \quad (8)$$

Here, *i* is the index of an input. Equation (6) implies that the synapses are connected by Bernoulli distributions with probability **p** = [$p_i$] (*i* = 0, …, *n*-1). If the synaptic connection is ON, an integer $c_i$ is assigned as its weight; otherwise, the synaptic weight (of an OFF connection) is set to 0. Equation (7) assures that the expectation of the synaptic weight in TrueNorth $E\{w_i'\}$ is equivalent to the weight of neural network $w_i$ represented by a real number. Similarly, equation (8) assures the expectation of input spikes is $x_i$ (0 ≤ $x_i$ ≤ 1), which corresponds to the normalized intensity (*e.g.* pixel value) of the $i^{th}$ input. Note that we omit the bias *b* because it can be included in $\mathbf{w} \cdot \mathbf{x}$ by adding an additional connection with weight *b* and a constant input 1. The same simplification can also be applied to the leak $\lambda$ in Equation (3).

As such, suppose **w'** and **x'** are independent, the expectation of the sum of the weighted inputs in TrueNorth $E\{y'\}$ can be derived by:

$$E\{y'\} = E\left\{\sum_{i=0}^{n-1} w_i' x_i'\right\} = \sum_{i=0}^{n-1} E\{w_i'\} E\{x_i'\}$$
$$= \sum_{i=0}^{n-1} p_i c_i x_i = \sum_{i=0}^{n-1} w_i x_i = y \quad (9)$$

indicating that $E\{y'\}$ is the same as *y* in the original neural network model.

According to the central limit theorem (CLT) [11][12], the distribution of *y'* can be approximated by a normal distribution. Assume the mean and the variance of *y'* are $\mu_{y'}$ and $\sigma_{y'}$, respectively. Both $\mu_{y'}$ and $\sigma_{y'}$ will be functions of **x'** parameterized by **p**. The cumulative probability function (CDF) of *y'* then becomes

$$P(y' \geq y_0) = \frac{1}{\sigma_{y'}\sqrt{2\pi}} \int_{-\infty}^{y_0} e^{-(x-\mu_{y'})^2/(2\sigma_{y'}^2)} dx = 1 - \frac{1}{2}\left[1 + erf(\frac{y_0 - \mu_{y'}}{\sqrt{2}\sigma_{y'}})\right], \quad (10)$$

where *erf* denotes the Gauss error function. Hence, the possibility for a neuron to spike (*i.e.* the expected output of the neuron) is:

$$E\{z'\} = P(y' \geq 0) = 1 - \frac{1}{2}\left[1 + erf(\frac{-\mu_{y'}}{\sqrt{2}\sigma_{y'}})\right]. \quad (11)$$

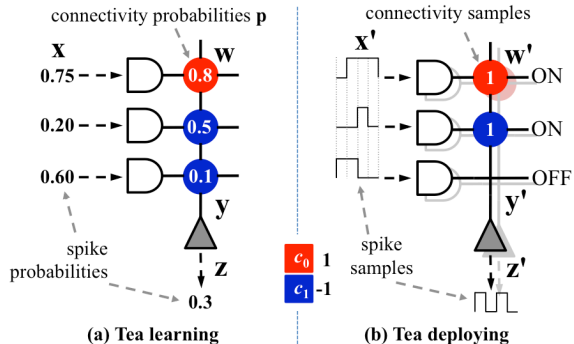

**Figure 2. An overview of the learning and deploying on TrueNorth.**

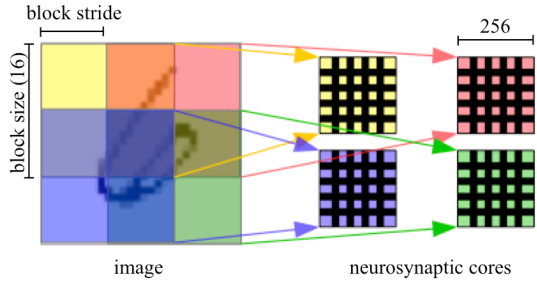

**Figure 3. A neural network with four neuro-synaptic cores.**

Equations (5)-(11) represents the process of deploying a neural network model with real number weights to TrueNorth: Firstly, we need to set the input spiking probability $P(x_i' = 1)$ according to the normalized intensity $x_i$; secondly, the activation $z$ of a hidden or output neuron is also translated into the predicted spiking probability of the neuron, such as $P(y' \geq 0)$; thirdly, the expectation of the synaptic weight $E\{w_i'\}$ shall be equivalent to the corresponding weight of neural network $w_i$; and lastly, Equation (11) works as the differentiable activation function. The optimization target becomes learning a synaptic connectivity probability **p** for the best inference accuracy.

Following Tea learning method, we trained a neural network illustrated in Figure 3 to perform handwritten digit recognition on MNIST database [13]. The network utilizes 4 neuro-synaptic cores, each of which receives a 16×16 (256 pixels) block from the image via its input neurons, with output axons from all neuro-synaptic cores being merged to 10 output classes for digit classification. Our experiment shows that after 10 training epochs over all the training samples, the network in Caffe (a CPU and GPU framework for deep learning [14]) with floating-point precision achieves a test accuracy of 95.27%. However, after deploying the same learned model to TrueNorth by sampling the connectivity of synapses with the learned connection probability **p**, the test accuracy quickly drops to 90.04% because of the aforementioned quantization induced accuracy loss. The recognition accuracy can be compensated back to 94.63% if instantiating 16 copies of the networks and averaging their results. However, this workaround requires total 64 neuro-synaptic cores for the presented example and is obviously not a scalable solution: the available neuro-synaptic cores on the TrueNorth chip will be quickly depleted as the scale of the implemented neural networks grows up.

### 3.2 Theoretical Analysis on Accuracy Loss

In this section, we will perform theoretical analysis to understand how the deployment on TrueNorth affects the performance and accuracy of a neural network. The obtained observation and the corresponding analysis constitute the foundation of our proposed probability-biased learning method.

As discussed in Section 2 and 3.1, current workaround to compensate accuracy loss is averaging the results over multiple copies instantiated in spatial and temporal domains. Thus, in this work, we mainly focus on mitigating the adverse impacts of the randomness of each copy, which actually are embodied in the sum of the weighted inputs $y'$. The accuracy loss is essentially determined by the deviation of $y'$ from $y$ in the original model, which is:

$$\Delta y = y' - y = \sum_{i=0}^{n-1} w_i' x_i' - \sum_{i=0}^{n-1} w_i x_i. \quad (12)$$

Ideally, the stochastic input spikes of Equation (8) and the connection probability function across the instantiations of networks in Equation (6) shall make the expectation of the deviation satisfy

$$E\{\Delta y\} = 0, \quad (13)$$

which indicates an unbiased learned model for TrueNorth. However, in the real implementation with the limited numbers of the input spikes and the instantiations of networks, the variance of the deviation $\Delta y$ inevitably causes accuracy degradation.

Assume that the events of the input spikes and the syntactical connections are independent, we then have the variance of $\Delta y$:

$$var\{\Delta y\} = \sum_{i=0}^{n-1} var\{w_i' x_i'\}. \quad (14)$$

This result leads to the conclusion that the variance of $\Delta y$ is incurred by both the synaptic randomness and the spiking randomness. In other words, minimizing the randomness helps increasing the certainty of $y'$ of the deployment on TrueNorth, and consequently, improving system accuracy and/or reducing the number of the needed spatial and temporal copies.

Note that $x_i'$ is solely determined by the external input data, not the deployment on TrueNorth. Only the network structure and the weights **w** affect the design efficacy. Hence, we attempt to minimize the synaptic variance $var\{w_i'\}$ as:

$$var\{w_i'\} = E\{(w_i')^2\} - E\{w_i'\}^2 = c_i^2 p_i (1 - p_i). \quad (15)$$

The largest $var\{w_i'\}$ occurs when the synapse has an equal ON and OFF probability, *i.e.*, $p_i = 0.5$. The positions of the minimum variances (*i.e.*, $p_i = 1$ and $p_i = 0$) are consistent to our intuition: the uncertainty can be eliminated if the synaptic connections are deterministically ON or OFF.

### 3.3 Proposed Learning Method

Based on the above observation and analysis, we propose a new learning method, which biases synaptic connectivity probability to deterministic boundaries (*i.e.*, $p_i = 1$ or $p_i = 0$) and therefore improves the inference accuracy of TrueNorth implementation.

Equations (6) and (7) indicate that for a given $c_i$, when training a TrueNorth model with floating-point precision, biasing the connectivity probability **p** is equivalent to tuning the weights **w**. Thus, the connectivity probability and the weights can be interchangeable in this context.

In the back propagation training of neural networks, an effective technique to enforce constraints on weights is to add weight

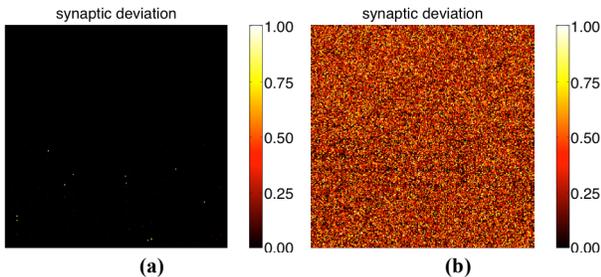

**Figure 4.** (a) is the synaptic weight deviation of deployment by probability-biased learning, comparing with that of Tea learning in (b).

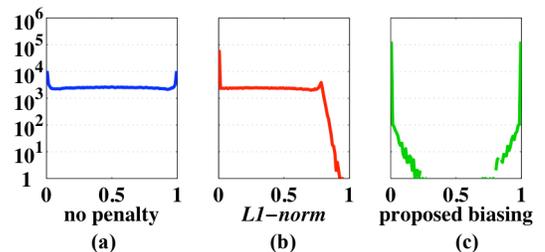

**Figure 5.** Probability (weight) distribution under different penalties.

penalty to the minimization target function [15], such as

$$\hat{E}(\mathbf{w}) = E_D(\mathbf{w}) + \lambda \cdot E_W(\mathbf{w}). \qquad (16)$$

Here $\mathbf{w}$ is the vector of all weights in the neural network; $E_D(\mathbf{w})$ is the inference error to be minimized; $E_W(\mathbf{w})$ is the weight penalty; and $\lambda$ is a regularization coefficient. An efficient penalty to bias $\mathbf{w}$ to zeros is *L1-norm*, which is $E_W(\mathbf{w}) = \|\mathbf{w}\| = \sum_{k=1}^{M} |w_k|$.

Our experiments on MNIST show that in all three layers of a feed-forward neural network in [16], 88.47%, 83.23% and 29.6% of weights can be zeroed out, with slight classification accuracy drop from 97.65% to 96.87%. The neural network has two hidden layers composed of 300 and 100 hidden neurons, respectively.

Although *L1-norm* is an effective method to bias weights to zeros, it doesn't help to reduce synaptic variance. For example, for the test bench of MNIST handwritten digit recognition, the histogram distributions of the connectivity probability (weight) without any penalty function and the one after applying *L1-norm* penalty are presented in Figure 5(a) and (b), respectively. The recognition accuracies of the two designs are very close (95.27% vs. 95.36%). Interestingly, the experiment shows that a large portion of the weights fall near the poles ($p_i = 1$ or 0) in Figure 5(a) even without any penalty. *L1-norm* can further bias the probability distribution to zero but pushes it away from the deterministic pole of $p_i = 1$, and the probability around the worst point ($p_i = 0.5$) is still kept at a relatively high level. As a result, after deploying the weights in Figure 5(b) on the four neuro-synaptic cores of TrueNorth shown in Figure 3, the accuracy degrades to 89.83%.

The above observation inspired us to propose a new probability-biased learning method that can aggressively bias the probability histogram toward the two poles (and away from the worst point at centroid) to improve the inference accuracy. In particular, the proposed scheme uses the following penalty function to bias the histogram toward the two poles:

$$E_b(\mathbf{w}) = \| |\mathbf{w} - a| - b \| = \sum_{k=1}^{M} | |w_k - a| - b |, \qquad (17)$$

where $a$ denotes the centroid it biases away and $b$ is the distance between the poles and the centroid. Note that if we represent entire $|\mathbf{w}-a|-b$ with variable $\mathbf{s}$, the biasing penalty $E_b(\mathbf{w})=\|\mathbf{s}\|$ becomes a *L1-norm* penalty, which tends to pull $\mathbf{s}$ toward zeros. Hence, the purpose of the biasing penalty function in Equation (17) is to pull $\mathbf{w}$ to $a + b$ and $a - b$. A special case is $a = b = 0.5$ that pushes connectivity probabilities to the poles at the two ends by enforcing the highest penalty on the worst case and the lowest (none) penalty on the best cases.

Figure 5(c) shows the learned probability histogram after applying our proposed biasing penalty function by setting $a = b = 0.5$. The obtained recognition accuracy is 95.03%, which is slightly lower than the 95.27% of the network without any penalty. Nevertheless, this new model reaches an accuracy of 92.78% after being deployed to TrueNorth with connectivity quantization. Comparing to the previously received accuracy of 90.04% (89.83%) without any penalty (with *L1-norm* penalty), our learning method raises the recognition accuracy by 2.5% (2.71%) using the same number of the synaptic cores. The improvement comes from the fact that almost all connective probabilities of synapses are biased to the deterministic states (0 or 1), as illustrated in Figure 5(c). It implies a minimized synaptic variance as depicted in Equation (15). More evaluations on the hardware resource reduction, performance, and accuracy that achieved by our learning method will be presented Section 4 in detail.

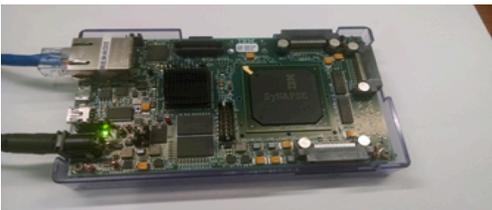

**Figure 6. NS1e development platform with the IBM TrueNorth chip.**

To illustrate the efficacy of our method, *e.g.*, the synaptic variance in neuro-synaptic cores, the maps of synaptic weight deviation between the desired learned TrueNorth model and the deployed TrueNorth model are extracted from IBM Neuro Synaptic Chip Simulator (NSCS) and plotted in Figure 4. In the figures, each deviation map corresponds to a randomly selected neuro-synaptic core in a TrueNorth chip, and each point in the map (totally 256x256 points) is the deviation of deployed synaptic weight from the desired weight with floating-point precision (note that the deviation is normalized by the maximum possible synaptic weight). Figure 4 (a) and (b) are the deviations obtained with biasing optimization and without any penalty optimization, respectively. Without applying our learning method, the deviation in Figure 4(b) is large, say, 24.01% of the synapses having deviations larger than 50%. After applying our probability-biased learning, 98.45% of the synapses have zero deviation, as shown in Figure 4(a). In fact, less than 0.02% of the synapses have deviations larger than 50%.

## 4. EXPERIMENTS

We use datasets in Table 1 for our classification experiments, including MNIST handwritten digits [13] and RS130 protein secondary structure dataset [17][18]. Protein secondary structure is the local conformation of the polypeptide chain and classified to three classes: alpha-helices, beta-sheets, and coil [18]. In this section, Section 4.1 introduces the hardware platform and *test bench 1*; Section 4.2, 4.3 and 4.4 analyze the accuracy, core occupation and performance efficiency based on *test bench 1*, respectively; Finally, Section 4.5 extends our experiments to more neural networks with different scales trained in different datasets, to demonstrate the adaptability and scalability of our method.

### 4.1 Platform and Test Bench

Our experiment is run on both the IBM Neuro Synaptic Chip Simulator and the NS1e development platform hardware, which contains one IBM TrueNorth chip (shown in Figure 6). We use MNIST dataset and the neural network in Figure 3 as the *test bench 1* for consistent analyses in Section 4.2, 4.3 and 4.4. For use of TrueNorth, pixel values are scaled to [0, 1] and converted to spikes. The network is trained in Caffe and deployed to TrueNorth by randomly sampling synaptic connections. The classification accuracy achieved with the network in Caffe is 95.27% without penalty, and 95.03% with biasing penalty, as already presented in Section 3.3.

In the following sections, we will evaluate efficiencies of our proposed learning method in accuracy, core occupation, and performance (speed). Note that, without specific mention, here accuracy refers to the test accuracy after deploying on TrueNorth.

### 4.2 Accuracy Efficiency

Firstly, we show the efficiency of our probability-biased learning method to retain the accuracy. Figure 7 illustrates resultant accuracies from various combinations of network (1 to 16 copies) and spike (1 to 4 spikes per frame) duplication. Here spike per frame (*spf*) is defined as the number of spike samples utilized to encode each pixel/activation. The red and yellow surface is the accuracy surface of Tea learning and our probability-biased learning, respectively. From Figure 7, we can see both spatial and temporal duplication can reduce accuracy loss caused by low quantitation precision. However, as duplications increase, the accuracy surfaces saturate to the plane limited by the original accuracy trained by Caffe (95%). Note that, we have averaged accuracy at each grid over ten results considering the randomness in TrueNorth.

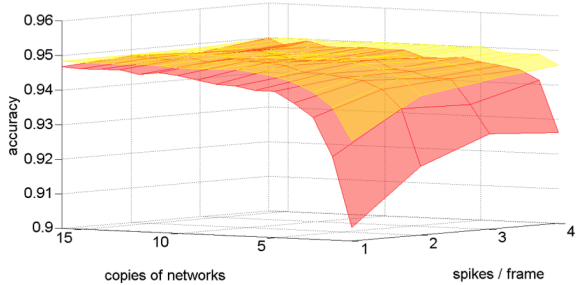

Figure 7. Absolute accuracies of Tea learning (red lower surface) and probability-biased learning (yellow upper surface).

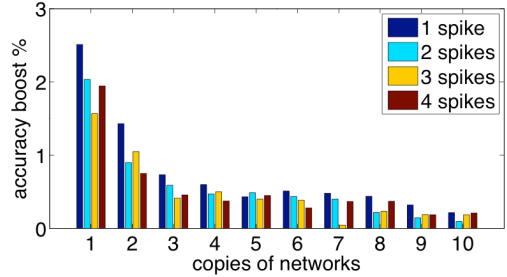

Figure 8. Accuracies boost (our method minus Tea learning).

We can see that our accuracy (yellow) surface covers above the original (red) one from Tea learning. It shows that under the same spatial and temporal duplications, our method retains better accuracy (especially when the number of duplication is small).

Figure 8 depicts the accuracy improvements achieved by our learning method w.r.t. Tea learning. The highest gain (2.5%) is reached at the lowest levels of duplication (one network copy and one *spf*). In general, when fewer copies of networks are occupied (which means more cores are saved), a higher accuracy improvement can be achieved, showing less reliance of our approach on spatial duplication.

### 4.3 Core Occupation Efficiency

To evaluate efficiency of our method in reducing resource consumption, we compare core occupation required for the same accuracy when utilizing our method versus Tea learning. Note that it is difficult to achieve exactly equivalent accuracy between two TrueNorth neural networks. Nonetheless, we show that even when adopting a comparison procedure that is biased toward Tea learning, our method still requires less cores. Be specific, the biased nature of our comparison occurs when the same accuracy cannot be obtained between the two compared methods – in such a case, we select the accuracy of the Tea learning as the baseline and compare it to the network required for the next greater level of accuracy when using our method. Table 2 (a) shows the comparison of the occupied cores between two tested TrueNorth neural networks and their corresponding accuracies. Accuracies are ordered in ascend and two types of neural networks with nearest accuracies are grouped (with gray shadow) in the table for easy comparison. In each group, the accuracy acheived by our method is either equal or higher than that of the current method. The number (and the percentage) of saved cores is recorded in the third row. These networks are denoted by N# and B#, where N/B indicates the model is learned without penalty (**N**one) or with **B**iasing penalty and # is the number of copies of the network.

Under 1 *spf* shown in Table 2 (a), our approach can save on average 49.5% cores with equivalent (or higher) accuracy. Moreover, the number of saved cores increases with the desired level of accuracy, showing the scalability of our method to higher inference accuracy, which is often the highest-priority demand for intelligent systems. Similar phenomena are observed when *spf* is larger. The average percentages of core reduction under different *spf* are depicted in Figure 9 (a). The core saving roughly increases with the *spf*, showing the adaptability and scalability of our method to different speed requirements (*spf*).

### 4.4 Performance Efficiency

We also evaluate the performance/speed efficiency of our learning method. The results are listed in Table 2 (b), which has a similar format to that of Table 2 (a) except that here # denotes the number of *spf*. With one copy of the neural network, our method, for example B2, can still achieve 0.60% higher accuracy with a speed of 2 *spf*, comparing to 13 *spf* (N13) using Tea learning method. This result can be translated to 6.5× speedup and similar results are also observed at different instantiated network copies.

Table 1. Test datasets

| Dataset | Description | Area | Training size | Testing size | Feature # | Class # |
|---|---|---|---|---|---|---|
| **MNIST** | Handwritten digits | Computer Engineering | 60,000 | 10,000 | 784 (28×28) | 10 |
| **RS130** | Protein secondary structure | Life Science | 17,766 | 6,621 | 357 | 3 |

Table 2. Core occupation and performance efficiency of probability-biased learning method

(a) Core occupation efficiency (1 *spf*)

| Accuracy | 0.904 | 0.924 | 0.929 | 0.935 | 0.938 | 0.939 | 0.942 | 0.942 | 0.943 | 0.944 | 0.945 | 0.946 | 0.947 | 0.947 |
|---|---|---|---|---|---|---|---|---|---|---|---|---|---|---|
| Network Copies | N1[*] | N2 | B1 | N3 | B2 | N4 | N5 | B3 | N7 | N9 | B4 | N10 | N16 | B5 |
| Saved Core | - | 4 (50.0%) | | 4 (33.3%) | | - | 8 (40.0%) | | - | 20 (55.6%) | | | 44 (68.8%) | |

(b) Performance efficiency (1 network copy)

| Accuracy | 0.904 | 0.920 | 0.927 | 0.928 | 0.929 | 0.932 | 0.933 | 0.934 | 0.940 | 0.943 | 0.946 | 0.947 | 0.948 | 0.950 |
|---|---|---|---|---|---|---|---|---|---|---|---|---|---|---|
| *spf* | N1[*] | N2 | N3 | N6 | B1 | N7 | N11 | N13 | B2 | B3 | B4 | B5 | B9 | B13 |
| Speedup | | | | 6 | | | | 6.5 | | | | | | |

[*] TrueNorth neural networks are denoted by N# and B#, where N/B indicates the model is learned without penalty (**N**one)/**B**iasing penalty. # is the number of network copies in (a) and *spf* in (b) respectively.

## 4.5 Adaptability and Scalability

We apply our learning method to five test benches to demonstrate its adaptability to different applications and scalability to different neural network structures. Table 3 lists various configurations of the test benches among which test bench 1 is the one used in previous discussions. Without losing generality, we adopt feed-forward neural network in our test benches. 357 one-dimensional features in RS130 dataset are reshaped to two dimensions (19×19) and sent to neuro-synaptic cores via similar structure in Figure 3. Output axons from all neuro-synaptic cores at the last hidden layer are merged to output classes for classification.

Figure 9(b) summarizes the core reductions achieved at different test benches. The benefit of our method varies with applications and network structures due to different randomness characteristics; Nonetheless, our method always substantially reduce the needed cores on TrueNorth over all test benches.

## 5. CONCLUSION

In this work, we theoretically analyze the impacts of low data precision in TrueNorth on inference accuracy, core occupation, and performance, and propose a probability-biased learning method to enhance the inference accuracy through reducing the random variance of each computation copy. Very encouraging results are obtained by implementing our new learning method on real TrueNorth chip: compared to the existing Tea learning method of TrueNorth, our method can achieve up to 68.8% reduction of the required neuro-synaptic cores or 6.5× speedup, with even slightly improved inference accuracy.

## 6. ACKNOWLEDGEMENT

This work was supported in part by AFRL FA8750-15-1-0176, NSF 1337198 and NSF 1253424. Any opinions, findings and conclusions or recommendations expressed in this material are those of the authors and do not necessarily reflect the views of AFRL, DARPA, NSF, or their contractors.

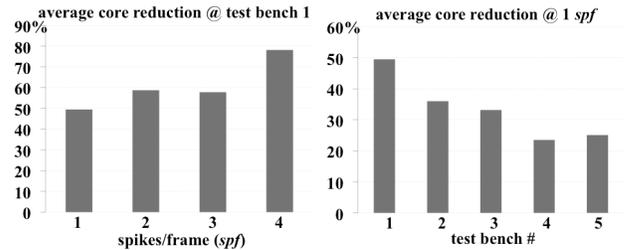

(a) Core efficiency vs. *spf*  (b) Core efficiency vs. test bench
**Figure 9. Adaptability and scalability of our method.**

**Table 3. Test Benches**

| Test bench | Dataset | Block stride | Hidden layer # | Cores per layer | Accuracy in Caffe |
|---|---|---|---|---|---|
| 1 | MNIST | 12 | 1 | 4 | 95.27% |
| 2 | MNIST | 4 | 1 | 16 | 96.71% |
| 3 | MNIST | 2 | 3 | 49~9~4 | 97.05% |
| 4 | RS130 | 3 | 1 | 4 | 69.09% |
| 5 | RS130 | 1 | 2 | 16~9* | 69.65% |

* 16 and 9 correspond to the cores utilized by 1st and 2nd hidden layer.